%% file: arxiv.tex
\documentclass[final]{cvpr}

\usepackage{times}
\usepackage{epsfig}
\usepackage{graphicx}
\usepackage{amsmath}
\usepackage{amssymb}
\usepackage{soul}
\usepackage{comment}
\usepackage{balance}

\usepackage{multirow}  
\usepackage{paralist}  
\usepackage{pifont}  

\newcommand{\xmark}{\ding{55}}

\usepackage[pagebackref=true,breaklinks=true,colorlinks,bookmarks=false]{hyperref}

\usepackage{xcolor}

\def\babel{BABEL}
\def\cmu{CMU MoCap}
\def\ntu{NTU RGB+D}
\def\ha12{HumanAct12}
\def\twoSAGCN{2s-AGCN}

\def\cvprPaperID{6843} 
\def\confYear{CVPR 2021}

\begin{document}

\title{BABEL: Bodies, Action and Behavior with English Labels}

\author{
    \textbf{Abhinanda R. Punnakkal}\thanks{\,Denotes equal contribution.}$\,\,^{,1}$  \qquad
    \textbf{Arjun Chandrasekaran}$^{*,1}$ \qquad 
    \textbf{Nikos Athanasiou}$^{1}$ \qquad \\
    \textbf{Alejandra Quirós-Ramírez}$^{2}$ \qquad 
    \textbf{Michael J. Black}$^{1}$ \qquad \\
    \small{$^1$Max Planck Institute for Intelligent Systems, T\"{u}bingen, Germany \quad
    $^2$Universit\"{a}t Konstanz, Konstanz, Germany} \\
    {\tt\small \{apunnakkal, achandrasekaran, nathanasiou, alejandra.quiros, black\}@tue.mpg.de} 
}

\maketitle

\begin{abstract}
Understanding the semantics of human movement -- the what, how and why of the movement -- is an important problem that requires datasets of human actions with semantic labels. 
Existing datasets take one of two approaches.
Large-scale video datasets contain many action labels but do not contain ground-truth 3D human motion.
Alternatively, motion-capture (mocap) datasets have precise body motions but are limited to a small number of actions. 
To address this, we present \babel, a large dataset with language labels describing the actions being performed in mocap sequences. 
\babel~labels about $43$ hours of mocap sequences from AMASS.
Action labels are at two levels of abstraction -- sequence labels which describe the overall action in the sequence, and frame labels which describe all actions in every frame of the sequence. 
Each frame label is precisely aligned with the duration of the corresponding action in the mocap sequence, and multiple actions can overlap. 
There are over $28k$ sequence labels, and $63k$ frame labels in BABEL, which belong to over $250$ unique action categories. 
Labels from \babel~can be leveraged for tasks like action recognition, temporal action localization, motion synthesis, etc. 
To demonstrate the value of \babel~as a benchmark, we evaluate the performance of models on 3D action recognition. 
We demonstrate that \babel~poses interesting learning challenges that are applicable to real-world scenarios, and can serve as a useful benchmark of progress in 3D action recognition. 
The dataset, baseline method, and evaluation code is made available, and supported for academic research purposes at \url{https://babel.is.tue.mpg.de/}.
\end{abstract}

\input{arxiv-sections/intro}
\input{arxiv-sections/rel-work}
\input{arxiv-sections/dataset}
\input{arxiv-sections/analysis}
\input{arxiv-sections/experiments}
\input{arxiv-sections/conclusion}

{\small
\bibliographystyle{ieee_fullname}
\balance
\bibliography{ref}
}
\end{document}

%% file: arxiv-sections/intro.tex
\section{Introduction}

A key goal in computer vision is to understand human movement in semantic terms. 
Relevant tasks include predicting semantic labels for a human movement, e.g., action recognition \cite{DBLP:journals/ivc/HerathHP17}, video description \cite{DBLP:conf/cvpr/XuMYR16}, temporal localization \cite{DBLP:conf/mir/SedmidubskyEZ19,DBLP:conf/iccv/Zhao0TY19}, 
and generating human movement that is conditioned on semantics, e.g., motion synthesis conditioned on actions \cite{DBLP:conf/mm/GuoZWZSDG020}, or sentences \cite{DBLP:conf/3dim/AhujaM19,lin20181}.

Large-scale datasets that capture variations in human movement and language descriptions that express the semantics of these movements, are critical to making progress on these challenging problems. 
Existing datasets contain detailed action descriptions for only 2D videos, e.g., ActivityNet \cite{DBLP:conf/mir/SedmidubskyEZ19}, AVA \cite{DBLP:conf/cvpr/GuSRVPLVTRSSM18} and HACS \cite{DBLP:conf/iccv/Zhao0TY19}. 
The large scale 3D datasets that contain action labels, e.g., \ntu~60 \cite{DBLP:conf/cvpr/ShahroudyLNW16} and \ntu~120 \cite{DBLP:journals/corr/abs-2004-14899} do not contain ground truth 3D human motion but only noisy estimates. 
On the other hand, motion-capture (mocap) datasets \cite{cmu_mocap,ghorbani2020movi,harvey2020robust,h36m_pami} are small in scale and are only sparsely labeled with very few actions. 
We address this shortcoming with \babel, a large dataset of diverse, densely annotated, actions with labels for all the actions  in a motion capture (mocap) sequence. 

\begin{figure}[t!]
    \centering
    \includegraphics[width=0.48\textwidth]{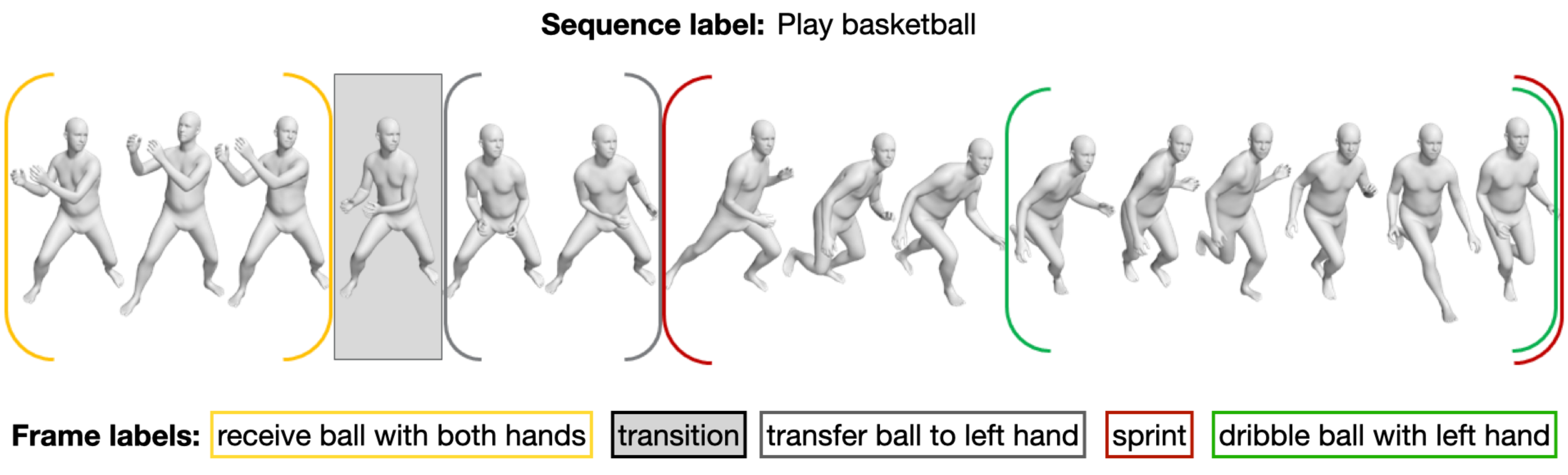}
    \caption{
        People moving naturally often perform multiple actions simultaneously, and sequentially, with transitions between them. 
    	\babel~contains sequence labels describing the overall action in the sequence, and frame labels where all frames and all actions are labeled. 
        Each frame label is precisely aligned with the frames representing the action (colored brackets). 
        This includes simultaneous actions (nested brackets) and transitions between actions (shaded gray box). 
	}
    \label{fig:teaser}
\end{figure}

We acquire action labels for sequences in \babel, at two different levels of resolution. 
Similar to existing mocap datasets, we collect a sequence label that describes the action being performed in the entire sequence, e.g., \texttt{Play basketball} in Fig.~\ref{fig:teaser}. 
At a finer-grained resolution, the frame labels describe the action being performed at each frame of the sequence, e.g., \texttt{transfer ball to the left hand}, \texttt{sprint}, etc. 
The frame labels are precisely aligned with the corresponding frames in the sequence that represent the action. 
\babel~also captures simultaneous actions, e.g., \texttt{sprint} and \texttt{dribble ball with left hand}. 
When collecting frame labels, we ensure that all frames in a sequence are labeled with at least one action, and all the actions in a frame are labeled. This results in dense action annotations for high-quality mocap data. 

\babel~leverages the recently introduced AMASS dataset \cite{DBLP:conf/iccv/MahmoodGTPB19} for mocap sequences. AMASS is a large corpus of mocap datasets that are unified with a common representation. 
It has $>43$ hours of mocap data performed by over $346$ subjects. 
The scale and diversity of AMASS presents an opportunity for data-driven learning of semantic representations for 3D human movement. 

Most existing large-scale datasets with action labels \cite{cmu_mocap,harvey2020robust,DBLP:journals/pami/LiuSPWDK20,DBLP:conf/icar/ManderyTDVA15,DBLP:conf/mir/SedmidubskyEZ19,DBLP:conf/cvpr/ShahroudyLNW16,DBLP:conf/iccv/Zhao0TY19,DBLP:journals/corr/abs-2004-14899} first determine a fixed set of actions that are of interest. 
Following this, actors performing these actions are captured (3D datasets), or videos containing the actions of interest are mined from the web (2D datasets). 
While this ensures the presence of the action of interest in the sequence, all other actions remain unlabeled. 
The sparse action label for a sequence, while useful, serves only as weak supervision for data-driven models that aim to correlate movements with semantic labels. This is suboptimal. 
3D datasets such as \ntu~\cite{DBLP:journals/pami/LiuSPWDK20,DBLP:conf/cvpr/ShahroudyLNW16} and \ha12~\cite{DBLP:journals/corr/abs-2004-14899} handle this shortcoming by cropping out segments that do not correspond to the action of interest from natural human movement sequences. 
While the action labels for the short segments are accurate, the cropped segments are unlike the natural, continuous human movements in the real-world. Thus, the pre-segmented movements are less suitable as training data for real-world applications. 

\begin{table*}[t!]
    \centering
    \begin{tabular}{c|c|c|c|c|c}
        \hline
        Dataset & GT motion? & \# Actions & \# Hours & Per-frame? & Continuous?\\
        \hline
        \cmu~\cite{cmu_mocap} &                         \checkmark &    23      & 9     & \xmark &      \checkmark \\
        MoVi \cite{ghorbani2020movi} &                  \checkmark &    20      & 9     & \xmark &      \checkmark \\
        Human3.6M \cite{h36m_pami} &                    \checkmark &    17      & 18    & \xmark &      \checkmark \\
        LaFan1 \cite{harvey2020robust} &                \checkmark &    12      & 4.6   & \xmark &      \checkmark \\
        \ha12~\cite{DBLP:conf/mm/GuoZWZSDG020} &        \xmark &        12      & 6     & \xmark &      \xmark \\
        \ntu~60 \cite{DBLP:conf/cvpr/ShahroudyLNW16} &  \xmark &        60      & 37    & \xmark &      \xmark \\
        \ntu~120 \cite{DBLP:journals/pami/LiuSPWDK20} &      \xmark &        120     & 74    & \xmark &      \xmark \\
        \hline
        \multirow{2}{*}{\babel~(ours)} & \multirow{2}{*}{\checkmark} & \multirow{2}{*}{260} & 43.5 & \xmark & \multirow{2}{*}{\checkmark}\\ 
                        & & & 37.5 & \checkmark & \\ 
        \hline
    \end{tabular}
    \vspace{5pt}    
    \caption{
        Comparison of existing datasets containing action labels for human movement. 
        \texttt{GT motion} indicates whether the human movements are accurate (mocap) or noisy estimates (e.g., via tracking). 
        \texttt{\#~Actions} indicates the total count of action categories in each dataset. 
        \texttt{\#~Hours} indicates the total duration of all sequences in the dataset. 
        \texttt{Per-Frame?} indicates whether the action labels are precisely aligned with the corresponding spans of movement in the sequence.  
        \texttt{Continuous?} indicates whether the movement sequences are original, continuous, human movements or short cropped segments containing specific actions. 
        \babel~uniquely provides large-scale dense (per-frame) action labels for 37.5 hours of natural, continuous, ground-truth human movement data from the AMASS \cite{DBLP:conf/iccv/MahmoodGTPB19} dataset. 
        In addition, \babel~provides a label that describes the overall action in the entire sequence, for 43.5 hours of mocap from AMASS. 
    }
    \label{tab:datasets}
\end{table*}

Our key idea with \babel~is that natural human movement often involves multiple actions and transitions between them. Thus, understanding the semantics of natural human movement not only involves modeling the relationship between an isolated action and its corresponding movement but also the relationship between different actions that occur simultaneously and sequentially. 
With \babel, our goal is to provide accurate data for statistical learning, which reflects the variety, concurrence and temporal compositions of actions in natural human movement. 

\babel~contains action annotations for about $43.5$ hours of mocap from AMASS, with $15472$ unique language labels. 
Via a semi-automatic process of semantic clustering followed by manual categorization, we organize these into $260$ action categories such as \texttt{greet}, \texttt{hop}, \texttt{scratch}, \texttt{dance}, \texttt{play instrument}, etc. 
The action categories in \babel~belong to 8 broad semantic categories involving simple actions (\texttt{throw}, \texttt{jump}), complex activities (\texttt{martial arts}, \texttt{dance}), body part interactions (\texttt{scratch}, \texttt{touch face}), etc. (see Sec.~\ref{subsec:label_processing}). 

\babel~contains a total of $28055$ sequence labels, and $63353$ frame labels. 
This corresponds to dense per-frame action annotations for $10892$ sequences ($>37$ hours of mocap), and sequence-level annotations for all $13220$ sequences ($>43$ hours of mocap). 
On average, a single mocap sequence has $6.06$ segments, with $4.02$ unique action categories. 
We collect the sequence labels via a web interface of our design, and the frame labels and alignments by adapting an existing web annotation tool, VIA \cite{dutta2019vgg} (see Sec.~\ref{subsec:data_collection}). 
Labeling was done by using Amazon Mechanical Turk \cite{amt}. 

We benchmark the performance of models on \babel~for the 3D action recognition task \cite{DBLP:conf/cvpr/ShahroudyLNW16}. 
The goal is to predict the action category, given a segment of mocap that corresponds to a single action span. 
Unlike existing datasets that are carefully constructed for the actions of interest, 
action recognition with \babel~more closely resembles real-world applications due to the long-tailed distribution of classes in \babel.
We demonstrate that \babel~presents interesting learning challenges for an existing action recognition model that performs well on \ntu~60. In addition to being a useful benchmark for action recognition, we believe that \babel~can be leveraged by the community for tasks like pose estimation, motion synthesis, temporal localization, few shot learning, etc. 

In this work, we make the following contributions:
(1) We provide the largest 3D dataset of dense action labels that are precisely aligned with their corresponding movement spans in the mocap sequence. 
(2) We categorize the raw language labels into over $250$ action classes that can be leveraged for tasks requiring categorical label sets such as 3D action recognition. 
(3) We analyze the actions occurring in \babel~sequences in detail, furthering our semantic understanding of mocap data that is already widely used in vision tasks. 
(4) We benchmark the performance of baseline 3D action recognition models on \babel, demonstrating that the distribution of actions that resembles real-world scenarios, poses interesting learning challenges. 
(5) The dataset, baseline models and evaluation code are publicly available for academic research purposes at \url{https://babel.is.tue.mpg.de/}.

%% file: arxiv-sections/rel-work.tex
\section{Related Work}
\label{sec:related_work}

\noindent
\paragraph{Language labels and 3D mocap data.} 
We first briefly review the action categories in large-scale 3D datasets, followed by a more detailed comparison in Table~\ref{tab:datasets}. 
The CMU Graphics Lab Motion Capture Database (CMU)~\cite{cmu_mocap} is widely used, and has 2605 sequences. The dataset has 6 semantic categories (e.g., `human interaction', `interaction with environment') that, overall, contain 23 subcategories, e.g., `two subjects', `playground', `pantomime'. 
Human3.6M \cite{h36m_pami} consists of 12 everyday actions in 6 semantic categories such as `walking variations' (`walking dog', `walking pair'), `full body upright variations' (`greeting', `posing'), etc. 
MoVi \cite{ghorbani2020movi} consists of everyday actions and sports movements e.g., `clapping hands', `pretending to take picture', etc. 
KIT Whole-Body Human Motion Database (KIT)~\cite{DBLP:conf/icar/ManderyTDVA15} focuses on both human movement and human-object interaction~\cite{DBLP:conf/eccv/TaheriGBT20}
containing grasping and manipulation actions in addition to activities such as climbing and playing sports. 
LaFan1~\cite{harvey2020robust} is a recent dataset containing 15 different actions, including locomotion on uneven terrain, free dancing, fight movements, etc. 
These characterize the movement in the entire mocap sequence via simple tags or keywords. 
In contrast, the KIT Motion-Language Dataset~\cite{DBLP:conf/icar/ManderyTDVA15} describes motion sequences with natural language sentences, e.g., `A person walks backward at a slow speed'. 
While our motivation to learn semantic representations of movement is similar, action labels in \babel~are precisely aligned with the sequence. 

\noindent
\paragraph{Frame actions labels in 3D mocap.}
The CMU MMAC dataset \cite{DBLP:conf/cvpr/SpriggsTH09} 
contains precise frame labels for a fixed set of 17 cooking actions (including `none'). 
Arikan et al.~\cite{DBLP:journals/tog/ArikanFO03} and Muller et al.~\cite{DBLP:conf/sca/MullerBS09} partially automate labeling temporal segments for mocap using action classifiers. 
While these works assume a known, fixed set of classes, in \babel, we identify and precisely label all actions that occur in each frame. 

\noindent
\paragraph{Action labels and tracked 3D data.} 
\ntu~60 \cite{DBLP:conf/cvpr/ShahroudyLNW16} and 120 \cite{DBLP:journals/pami/LiuSPWDK20} are large, widely used datasets for 3D action recognition. 
In \ntu, RGBD sequences are captured via 3 Kinect sensors which track joint positions of the human skeleton. 
\ntu~has segmented sequences corresponding to specific actions. There are 3 semantic categories -- `Daily actions' (`drink water', `taking a selfie'), `Medical conditions' (`sneeze', `falling down') and `Mutual actions' (`hugging', `cheers and drink'). 
These datasets contain short cropped segments of actions, which differ from \babel~sequences, which are continuous, reflecting natural human movement data. 
The ability to model actions that can occur simultaneously, sequentially and the transitions between them is important for application to real world data~\cite{DBLP:conf/mir/SedmidubskyEZ19}. 
See Table~\ref{tab:datasets} for further comparison. 

\noindent
\paragraph{2D temporal localization}. 
Many works over the years have contributed to progress in the action localization task \cite{DBLP:journals/cviu/IdreesZJGLSS17,DBLP:conf/eccv/SigurdssonVWFLG16,yeung2018every}. 
ActivityNet \cite{DBLP:conf/mir/SedmidubskyEZ19} contains 648 hours of videos and 200 human activities that are relevant to daily life, organized under a rich semantic taxonomy. 
It has 19,994 (untrimmed) videos, with an average of 1.54 activities per video. 
More recently, HACS \cite{DBLP:conf/iccv/Zhao0TY19} provides a larger temporal localization dataset with 140,000 segments of actions that are cropped from 50,000 videos that span over 200 actions. 
AVA~\cite{DBLP:conf/cvpr/GuSRVPLVTRSSM18} is another recent large-scale dataset that consists of dense annotations for long video sequences for 80 atomic classes. In~\cite{Wu2015WatchnpatchUU}, the authors introduce a test recorded by Kinect v2 in which they describe activities as compositions of action interactions with different objects.
While \babel~also contains temporally annotated labels, it does not assume a fixed set of actions that are of interest. On the other hand, with \babel, we 
 elicit labels for all actions in the sequence including high-level (`eating'), and low-level actions (`raise right hand to mouth'). 

%% file: arxiv-sections/dataset.tex
\section{Dataset}

We first provide details regarding the crowdsourced data collection process. We then describe the types of labels in \babel, and the label processing procedure. 

\subsection{Data collection}
\label{subsec:data_collection}

We collect \babel~by showing rendered videos of mocap sequences from AMASS \cite{DBLP:conf/iccv/MahmoodGTPB19} to human annotators and eliciting action labels (Fig.~\ref{fig:stage2_interface}). 
The mocap is processed to make sure the person in the video faces the annotator in the first frame. We observe that a sequence labeled as \texttt{pick up object} often also involves other actions such as walking to the object, bending down to pick up the object, grasping the object, straightening back up, turning around and walking away. 
We argue that labeling the entire sequence with the single label is imprecise, and problematic. 
First, many actions such as \texttt{turn} and \texttt{grasp} are ignored and remain unlabeled although they may be of interest to researchers \cite{DBLP:conf/eccv/TaheriGBT20}. 
Second, sequence labels provide weak supervision to statistical models, which are trained to map the concept of \texttt{picking up object} to the whole sequence when it, in fact, contains many different actions. 
To illustrate this point, we examine a typical sequence (see Qualitative Example 1 in the project website), and find that only $20\%$ of the duration of the sequence labeled as \texttt{pick up and place object} corresponds to this action. Crucially, walking towards and away from the object -- actions that remain unlabeled -- account for $40\%$ of the duration. 
While this makes semantic sense to a human -- picking up and placing an object is the only action that changes the state of the world and hence worth mentioning, this might be suboptimal training data to a statistical model, especially when the dataset also contains the confusing classes \texttt{walk}, \texttt{turn}, etc. 
Finally, using noisy labels as ground truth during evaluation does not accurately reflect the capabilities of models. 

We address this with action labels at two levels of resolution -- a label describing the overall action in the entire sequence, and fine-grained labels that are aligned with their corresponding spans of movement in the mocap sequence.

\subsection{\babel~action labels}

We collect \babel~labels in a two-stage process -- first, we collect sequence labels, and determine whether the sequence contains multiple actions. 
We then collect frame labels for the sequences where $2$ annotators agree that there are multiple actions. 

\noindent
\paragraph{Sequence labels.}
In this labeling task, annotators answer two questions regarding a sequence. 
We first ask annotators if the video contains more than one action (yes/no).\footnote{Note that the initial `T-pose' for calibration, followed by standing are considered separate actions with a transition between them.} 
If the annotator chooses `no', we ask them to name the action in the video. 
If they instead choose `yes', we elicit a sequence label with the question, ``If you had to describe the whole sequence as one action, what would it be?'' 
We provide the web-based task interface in the project website. 

We ask annotators to enter the sequence labels in a text-box, with the option of choosing from an auto-complete drop-down menu that is populated with a list of basic actions. 
We specifically elicit free-form labels (as opposed to a fixed list of categories) from annotators to discover the diversity in actions in the mocap sequences. 
We find that in most cases, annotators tend to enter their own action labels. 
This also presents a challenge, acting as a source of label variance. Apart from varying vocabulary, free-form descriptions are subject to ambiguity regarding the `correct' level in the hierarchy of actions \cite{DBLP:conf/cvpr/GuSRVPLVTRSSM18}, e.g., \texttt{raise left leg}, \texttt{step}, \texttt{walk}, \texttt{walk backwards}, \texttt{walk backwards stylishly}, etc. 

We collect 2 labels per sequence, and in case of disagreement regarding multiple actions, a third label. 
We determine that a sequence contains a single action or multiple actions based on the majority vote of annotators' labels. 
Overall, \babel~contains $28055$ sequence labels\footnote{Note that a few sequences have additional labels.}. 

\begin{figure}[t!]
    \centering
    \includegraphics[width=0.48\textwidth]{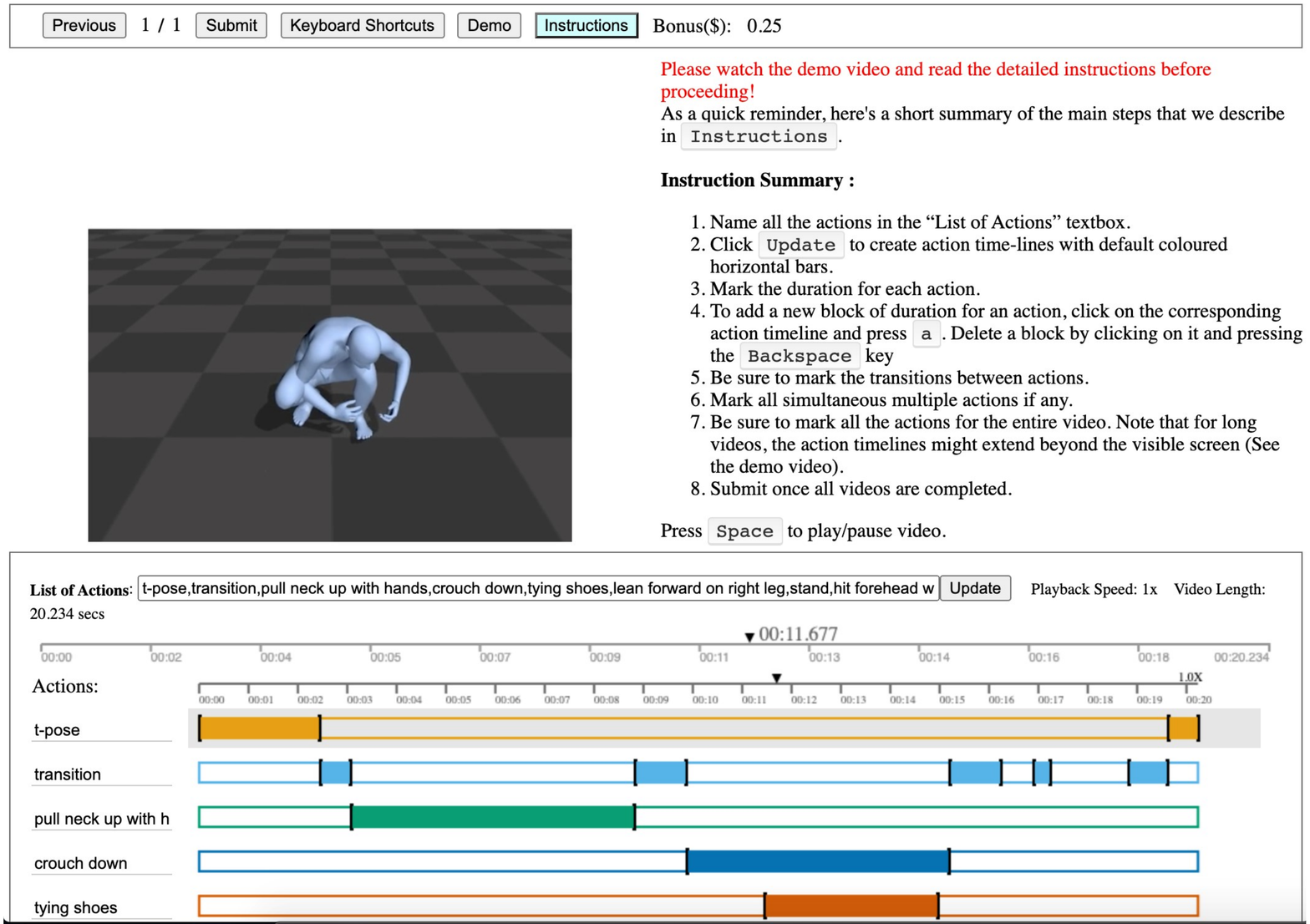}
    \caption{
        \babel~annotation interface to collect frame-level action labels. 
        Annotators first name all the actions in the video. 
        They then, precisely align the length of the action segment (colored horizontal bar) with the corresponding duration of the action in the video. 
        This provides dense action labels for the entire sequence. 
    }
    \label{fig:stage2_interface}
\end{figure}
 
\noindent
\paragraph{Frame labels.}
\label{sec:stage_2_interface}
Frame labels contain language descriptions of all actions that occur in the sequence, and precisely identify the span in the sequence that corresponds to the action. 
We leverage an existing video annotation tool, VIA~\cite{dutta2019vgg}, and modify the front-end interface and back-end functionality to suit our annotation purposes. 
For instance, we ensure that every frame in the sequence is annotated with at least one action label. This includes `transition' which indicates a transition between two actions, or `unknown' which indicates that the annotator is unclear as to what action is being performed. 
This provides us with dense annotations of action labels for the sequence. 
A screenshot of the AMT task interface for frame label annotation in \babel~is shown in Fig.~\ref{fig:stage2_interface}. 

To provide frame labels, an annotator first watches the whole video and enters all the actions in the `List of Actions' text-box below the video. 
This populates a set of empty colored box outlines corresponding to each action. 
The annotator then labels the span of an action by creating a segment (colored rectangular box) with a button press. 
The duration of the segment and the start/end times can be changed via simple click-and-drag operations. 
The video frame is continuously updated to the appropriate time-stamp corresponding to the end time of the current active segment. This provides the annotator real-time feedback regarding the exact starting point of the action. 
Once the segment is placed, its precision can be verified by a `play segment' option that plays the video span corresponding to the current segment. In case of errors, the segment can be further adjusted. 
We provide detailed instructions via text, and a video tutorial that explains the task with examples, and demonstrates operation of the annotation interface. 
The web interface of the task is provided in the project website. 

We collect frame labels for $6663$ sequences where both annotators who provided sequence labels agree that the sequence contains multiple actions\footnote{Note that some sequences are labeled by more than 1 annotator.}.

Overall, \babel~contains dense annotations for a total of $66018$ action segments for $10892$ sequences. This includes both frame labels from sequences containing multiple actions, and sequence labels from sequences containing a single action. If an entire sequence has only a single action, it counts as 1 segment. 

\subsection{Annotators} 
We recruit all annotators for our tasks via the Amazon Mechanical Turk (AMT)\footnote{\url{https://www.mturk.com/}}
crowd-sourcing platform. 
These annotators are located either in the US or Canada. 
In the sequence label annotation task, we recruit $> 850$ unique annotators with $>5000$ HITs approved and an approval rate $> 95\%$. 
In the frame labeling task, which is more involved, we first run small-scale tasks to recruit annotators. For further tasks, we only qualify about $130$ annotators who demonstrate an understanding of the task and  provide satisfactory action labels and precise segments in the sequence. 
In both tasks, we plan for a median hourly pay of $\sim \$12$. 
We also provide bonus pay as an incentive for thorough work in the frame labeling task (details in Sup.~Mat.). 

\begin{figure*}[t!]
    \centering
    \includegraphics[width=0.9\textwidth]{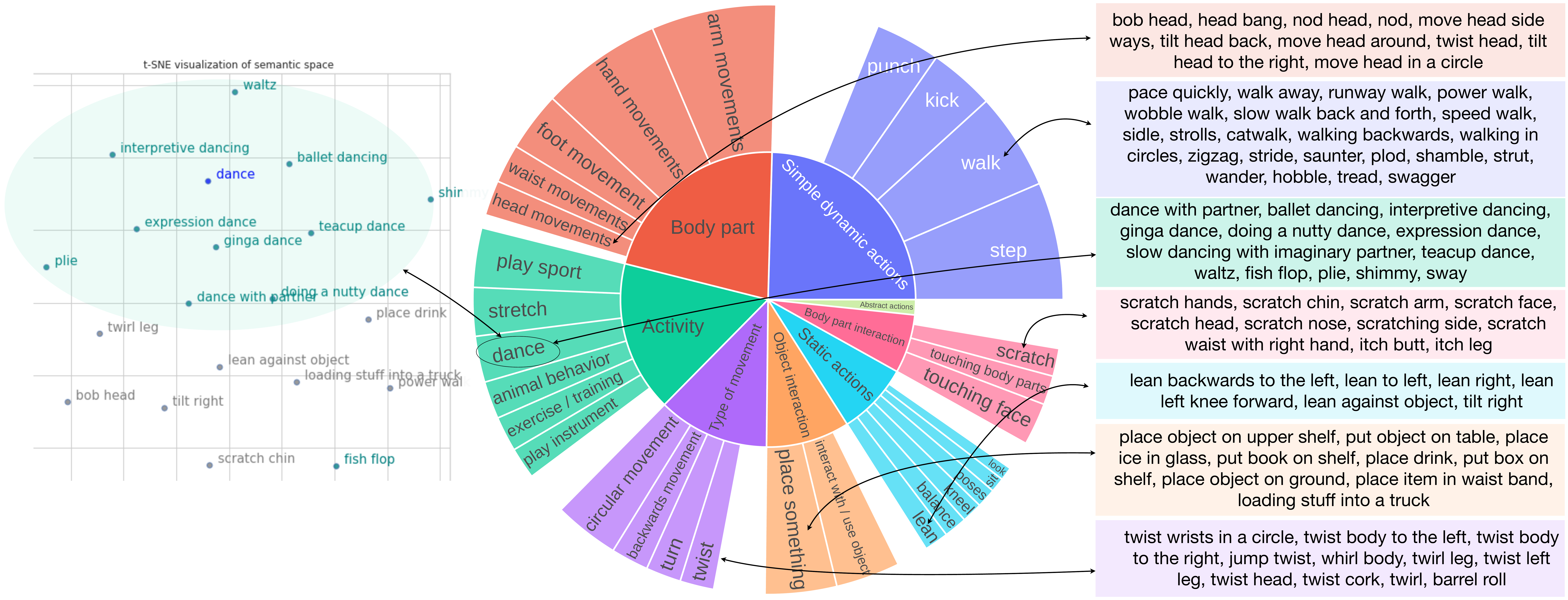}    
    \caption{
    \emph{Left.} 2D t-SNE \cite{maaten2008visualizing} visualization of the semantic space that we project raw labels into. Similar labels are grouped via K-means clustering. 
    The green shading and points represent the `dance' cluster and its members respectively. 
    \emph{Center.} Distribution of (a subset of) action categories (outer circle) under each semantic category (inner circle) in \babel. 
    The angle occupied by the action category is proportional to the number of unique raw label strings associated with it. 
    Action categories with a large number of fine-grained descriptions are shown. 
    \emph{Right.} Subset of the fine-grained descriptions associated with selected action categories. 
    }
    \label{fig:sunburst}
\end{figure*}

\subsection{Label processing}
\label{subsec:label_processing}
 
\babel~contains a total of $15472$ unique raw action labels. 
Note that while each action label is a unique string, labels are often semantically similar (\texttt{walk}, \texttt{stroll}, etc.), are minor variations of an action word (\texttt{walking}, \texttt{walked}, etc.) or are misspelled. 
Further, tasks like classification require a smaller categorical label set. 
We organize the raw labels into two smaller sets of semantically higher-level labels -- action categories, and semantic categories. 

\noindent
\paragraph{Action categories.} 
We map the variants of an action into a single category via a semi-automatic process that involves clustering the raw labels, followed by manual adjustment. 

We first pre-process the raw string labels by lower-casing, removing the beginning and ending white-spaces, and lemmatization. 
We then obtain semantic representations for the raw labels by projecting them into a 300D space via Word2Vec embeddings~\cite{w2vec}. 
Word2Vec is a widely used word embedding model that is based on the distributional hypothesis -- words with similar meanings have similar contexts. Given a word, the model is trained to predict surrounding words (context). An intermediate representation from the model serves as a word embedding for the given word. 
For labels with multiple words, the overall representation is the mean of the Word2Vec embeddings of all words in the label. Labels containing words that are semantically similar, are close in the representation space. 

We cluster labels that are similar in the representation space via K-means ($K=200$ clusters). 
This results in several semantically meaningful clusters, e.g., \texttt{walk}, \texttt{stroll}, \texttt{stride}, etc. which are all mapped to the same cluster. 
We then manually verify the cluster assignments and fix them to create a semantically meaningful organization of the action labels. 
Raw labels that are not represented by Word2Vec (e.g., \texttt{T-pose}) are manually organized into relevant categories in this stage. 
For each cluster, we determine a category name that is either a synonym (`walk' $\leftarrow$ \{\texttt{walk}, \texttt{stroll}, \texttt{stride}\}) or hypernym (`walk' $\leftarrow$ \{\texttt{walk forward}, \texttt{walk around}\}) that describes all action labels in the cluster. 

Some raw labels, e.g., \texttt{rotate wrists} can be composed into multiple actions like \texttt{circular movement} and \texttt{wrist movement}. 
Thus, raw labels are occasionally assigned membership to multiple action categories. 

Overall, the current version of \babel~has $260$ action categories. 
Interestingly, the most frequent action in \babel~is `transition' -- a movement that usually remains unlabeled in most datasets. There are $18447$ transitions between different actions in \babel. 
Unsurprisingly, the frequency of actions decreases exponentially following Zipf's law -- the $50$th most frequent action category \texttt{catch} occurs $417$ times, the $100$th most frequent action category \texttt{misc. activities}, occurs $86$ times, and the $200$th most frequent action category \texttt{disagree}, occurs $8$ times. 
We visualize the action categories containing the largest number of raw labels (cluster elements) in Fig.~\ref{fig:sunburst} (outer circle). 
Raw labels corresponding to these categories are shown on the right. 
We provide histograms of duration and number of segments per-action, in the Sup.~Mat and project webpage. 

\noindent
\paragraph{Semantic categories of labels.}
Action categories often reflect qualitatively different types of actions like interacting with objects, 
actions that describe the trajectory of movement,
complex activities involving multiple actions, etc. 
We formalize the different types of actions in \babel~into $8$ semantic categories (inner circle in Fig.~\ref{fig:sunburst}): 
\begin{compactenum}
    \item \textbf{Simple dynamic actions} contain low-level atomic actions -- \texttt{walk}, \texttt{run}, \texttt{kick}, \texttt{punch}, etc.
    \item \textbf{Static actions} involve transitioning to and/or maintaining a certain posture -- \texttt{sit}, \texttt{stand}, \texttt{kneel}, etc. 
    \item \textbf{Object interaction:} e.g., \texttt{place something}, \texttt{move something}, \texttt{use object}, etc. 
    \item  \textbf{Body part interaction} contains actions like \texttt{touching} \texttt{face}, \texttt{scratch}, etc.~which typically involve self-contact of body parts. 
    \item \textbf{Body part} describes the movement of a specific body part -- \texttt{raise arm}, \texttt{lower head}, \texttt{rotate wrist}, etc. 
    \item \textbf{Type of movement} contains actions that describe the trajectory of movement of either a body part or the whole body -- \texttt{twist}, \texttt{circular movement}, etc.  
    \item \textbf{Activity} contains complex actions that often involve multiple low-level actions -- \texttt{play sports=\{run, jump\}}, \texttt{dance=\{stretch, bend\}}, etc. 
    \item \textbf{Abstract actions} contain actions which often refer to the emotional state of the person and whose physical realizations could have large variance -- \texttt{excite}, \texttt{endure}, \texttt{learn}, \texttt{find}, etc. There are only a few abstract actions in \babel. 
\end{compactenum}

The diversity in the types of action labels in \babel~can be leveraged by tasks modeling movement at various levels of semantic abstraction, e.g., movement of body parts like \texttt{circular movement of wrist} at a low level, or high-level semantic activities such as \texttt{dancing the waltz}. 
Further, depending on the task and model, one can exploit either the discrete set of action categories (e.g., action recognition), or embed the raw action labels into a semantic space to provide a semantic representation of the segment of movement (e.g., action synthesis). 

We provide the full set of semantic categories, action categories, and raw action labels in \babel~in the project webpage.

%% file: arxiv-sections/analysis.tex
\section{Analysis}

Natural human movement often contains multiple actions and transitions between them. 
Modeling the likelihood of simultaneous actions and action transitions has applications in reasoning about action affordances in robotics and virtual avatars, motion synthesis \cite{DBLP:journals/corr/abs-1912-06079}, activity forecasting \cite{DBLP:conf/eccv/KitaniZBH12}, animation \cite{starke2019neural}, and action recognition.

\noindent
\subsection{Simultaneous actions}
Although people often perform multiple actions simultaneously in real life, this is rarely captured in labeled datasets. 
Recall from Sec.~\ref{sec:stage_2_interface} that in \babel, we ask annotators to label all actions that are occurring in each frame of the sequence. 
Overall, \babel~has $49952$ instances of simultaneous actions that occur with $2907$ unique pairs of action categories. 
Simultaneous actions are defined as actions that overlap for a duration of $> 0.1$ seconds. 
We exclude the overlap of an action with \texttt{transition} since this implies adjacent actions. 

Simultaneous actions often exhibit relationships such as: 
\begin{compactenum}
\item \textbf{Hierarchical}. Some simultaneous actions reflect the hierarchical structure in actions. For instance, a complex activity \& action comprising the activity, e.g., \texttt{eating food} \& \texttt{raise right hand to mouth}, and \texttt{dancing} \& \texttt{extend arms}. 
\item \textbf{Complementary}. The two actions are independent, e.g., \texttt{hold with left hand} \& \texttt{look right}. 
\item \textbf{Superimposed}. An action can move a certain body part that partly modifies another action; e.g., \texttt{carry with right hand} modifies the complex activity \texttt{walk}, and \texttt{right high kick} partly modifies the static (full body) action \texttt{fight stance}. 
\item \textbf{Compositional}. Actions involving the same body parts that result in a body or part movement that is a function of both actions, e.g., \texttt{walking} \& \texttt{turn}. 
\end{compactenum}

\noindent
\subsection{Temporally adjacent actions}
The dense labels in \babel~capture the progression of actions in mocap sequences. 
We analyze adjacent actions where where action $a_i$ follows $a_j$ (denoted by $a_j \rightarrow a_i$). 
$a_i$ and $a_j$ denote action segments, i.e., a contiguous set of frames corresponding to an action (and not the action for a single frame). Thus, $a_i \neq a_j$ if the actions are adjacent. 
We say $a_j \rightarrow a_i$ if the frame succeeding the last frame of $a_j$ is the first frame of $a_i$. 
In practice, we account for imprecise human temporal annotations by ignoring a small overlap in duration ($<0.1$ sec.) between $a_i$ and $a_j$. 
We also disregard the separation of actions by \texttt{transition};
i.e., $a_j \rightarrow a_i$ if $a_j \rightarrow a_t$ and $a_t \rightarrow a_i$, where $a_t = $ \texttt{transition}. 

\begin{figure}[t!]
    \centering
    \includegraphics[width=0.4\textwidth]{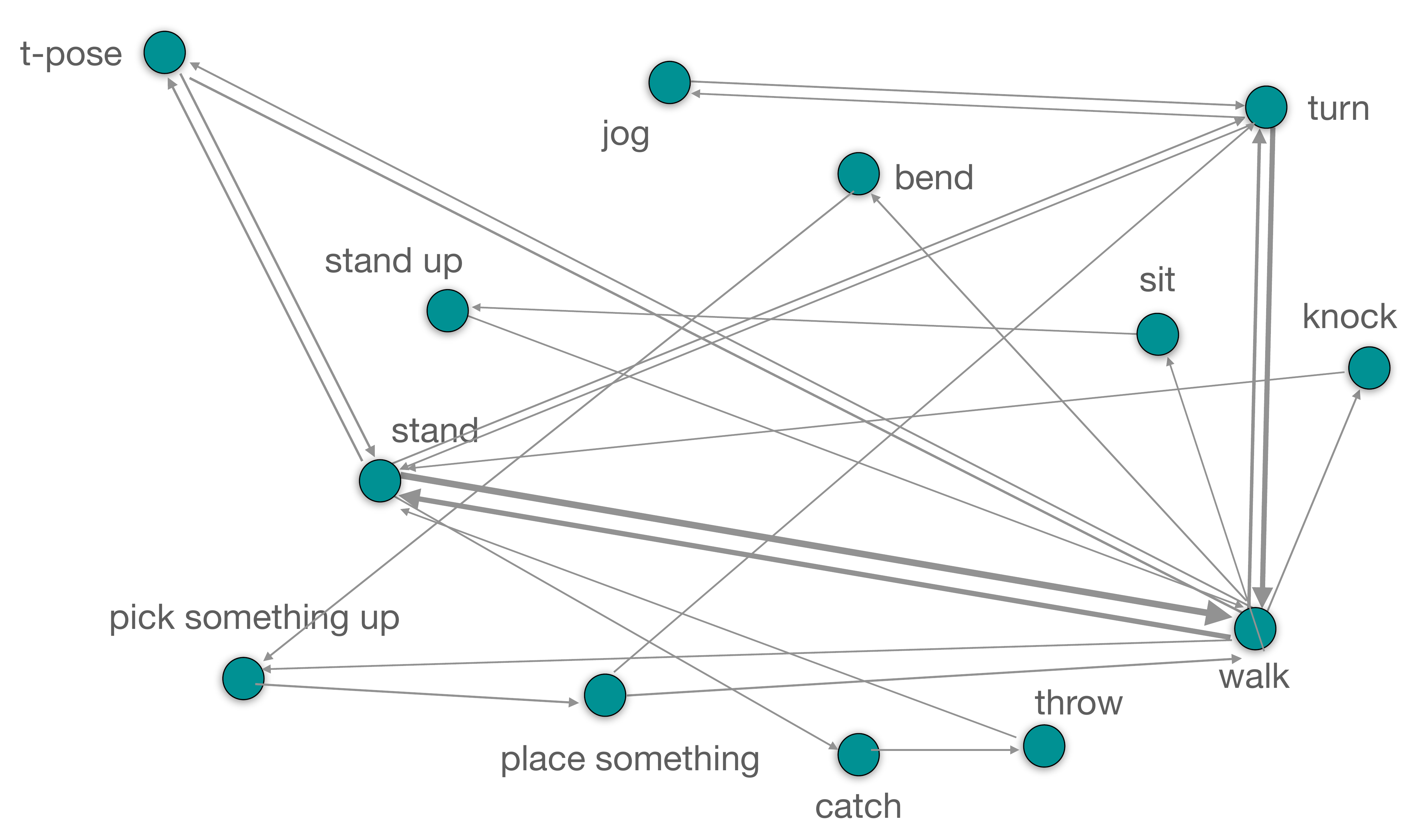}
    \caption{
        Node represent actions, and an edge represents a transition between these actions in the mocap sequence. 
        Edge thickness $\propto$ \texttt{Count}$(a_i \rightarrow a_j)$ (frequency of transition) in \babel. 
    }
    \label{fig:state_machine}
\end{figure}

We visualize the frequent transitions between actions, i.e., $a_j \rightarrow a_i$ sorted by \texttt{Count}($a_j \rightarrow a_i)$ in \babel, in Fig.~\ref{fig:state_machine}. 
We observe that \texttt{walk}, unsurprisingly, has the most diverse set of adjacent actions, i.e., \texttt{Count(walk}$\rightarrow a_i$\texttt{)} and \texttt{Count(}$a_j\rightarrow$\texttt{walk)} are large. 
While transitions between action pairs such as (\texttt{jog}, \texttt{turn}), (\texttt{walk}, \texttt{t-pose}) are bidirectional (with $\sim$ equal frequency), others have fewer adjacent actions. 
Some action categories with few transitions illustrate semantically meaningful action chains, e.g., \texttt{sit} $\rightarrow$ \texttt{stand up} $\rightarrow$ \texttt{walk} and \texttt{walk} $\rightarrow$ \texttt{bend} $\rightarrow$ \texttt{pick something up} $\rightarrow$ \texttt{place something}. 
Unidirectional transitions such as \texttt{sit} $\rightarrow$ \texttt{stand up} and \texttt{walk} $\rightarrow$ \texttt{sit} implicitly indicate the arrow of time \cite{DBLP:conf/cvpr/PickupPWSZZSF14}. 
Interestingly, the transition from \texttt{sit} $\rightarrow$ \texttt{stand up}, and the lack of transition from \texttt{sit} $\rightarrow$ \texttt{stand} delineates the subtle difference between the labels \texttt{stand} (static action of `maintaining an upright position') and \texttt{stand up} (dynamic action of `rising into an upright position'). 

Given the temporally adjacent actions in \babel, we attempt to model the transition probabilities between actions, i.e., $P(a_i|a_j)$. 
Concretely, we compute $P(a_i|a_{i-1}, a_{i-2}, a_{i-3})$, an order $3$ Markov Chain \cite{gagniuc2017markov}, 
and observe in Table~\ref{tab:mm_randomwalk} that random walks along this chain generate plausible action sequences for human movement. 

\begin{table}[t]
    \centering
    \begin{tabular}{c|p{7cm}}
        \hline
         \# & Transition of actions \\
        \hline
        1 & walk, transition, pick up, set down, transition, walk clockwise, transition, stand \\ 
        2 & a-pose, transition, wave hands in and out, wave arms in front of left, transition, cross left leg in circle gesture series, transition, t-pose \\ 
        3 &  looking left, standing, transition, looking right, standing \\
        4 & stepping forward, standing, turning back, walking back, walking forward, standing, losing balance, transition, turning around, walking, standing \\
        5 & step back, stand, transition, walk to the left \\         
        \hline
    \end{tabular}
    \vspace{5pt}    
    \caption{
        Random walk samples based on action transition probabilities learned from \babel. The generated samples are plausible action sequences simulating natural human movement. 
    }
    \label{tab:mm_randomwalk}
\end{table}

\subsection{Bias}
There are a few potential sources of bias in \babel, which we report in the Sup.~Mat. 
Specifically, we discuss potential biases introduced by the interface design, pay structure, and label processing method. 
We also analyze the inter-annotator variation in \babel~labels by collecting $5$ unique annotations for each of $29$ sequences. 
We find that for the same sequence, annotators vary in the labeled action categories, the number of actions, and segments. In general, the variance appears to be larger for sequences of longer duration. We provide further details in the Sup.~Mat. 

%% file: arxiv-sections/experiments.tex
\section{Experiments}

The dense action labels in \babel~can be leveraged for multiple vision tasks like pose estimation, motion synthesis, temporal localization, etc. In this section, we demonstrate the value of \babel~for the 3D action recognition task \cite{DBLP:journals/pami/LiuSPWDK20,DBLP:conf/cvpr/ShahroudyLNW16}, 
where the goal is to predict a single action category $y \in \mathcal{Y}$, for a given motion segment $(\mathbf{x}_t, \cdots, \mathbf{x}_{t'})$. 

\paragraph{Motion representation.} 
A mocap sequence in AMASS, is an array of poses over time, $\mathbf{M} = (\mathbf{p}_1, \cdots, \mathbf{p}_L)$, where $\mathbf{p}_i$ are  pose parameters of the SMPL-H body model \cite{MANO:SIGGRAPHASIA:2017}. 

For consistency with prior work, we predict the $25$-joint skeleton used in \ntu~\cite{DBLP:conf/cvpr/ShahroudyLNW16}
from the vertices of the SMPL-H mesh; see Sup.~Mat.

Thus, we represent a  movement sequence as $\mathbf{X} = (\mathbf{x}_1, \cdots, \mathbf{x}_L)$, where $\mathbf{x}_i \in \mathbb{R}^{J \times 3}$ represents the position of the $J(=25)$ joints in the skeleton, in Cartesian co-ordinates, $(x, y, z)$. 

\noindent
\paragraph{Labels.} 
In \babel, a raw action label is mapped to the segment of human movement $\mathbf{X}_s = (\mathbf{x}_{ts}, \cdots, \mathbf{x}_{te})$ corresponding to the action. 
Recall that a raw action label for a segment can belong to multiple action categories $\mathcal{Y}_s$ (e.g., \texttt{rotate wrists} $\rightarrow$ \texttt{circular movement}, \texttt{wrist movement}). 
Overall, \babel~contains $N$ movement segments, and their action categories $(\mathbf{X}_s, \mathcal{Y}_s)^N$.

\noindent
\paragraph{Architecture.} 
We benchmark performance on \babel~with the 2-Stream-Adaptive Graph Convolutional Network (\twoSAGCN) \cite{shi2019two}, a popular architecture that performs graph convolutions spatially (along bones in the skeleton), and temporally (joints across time). 
Crucially, the graph structure follows the kinematic chain of the skeleton in the first layer but is adaptive -- the topology is a function of the layer and the sample. 
The model achieves good performance on both 2D and 3D action recognition. 
GCNs remain the architecture of choice even in more recent state-of-the-art approaches \cite{DBLP:conf/cvpr/Cheng0HC0L20}. 

\twoSAGCN~consists of two streams with the same architecture -- one which accepts joint positions, and the other, bone lengths and orientations, as input respectively. The final prediction is the average score from the two streams. 
On \ntu, \twoSAGCN\/ achieves achieves an accuracy of $88.5\%$ on the cross-subject task. 
In our experiments, we use only the joint stream, which achieves $2\%$ lower accuracy compared to \twoSAGCN~\cite{shi2019two}. 

\noindent
\paragraph{Data pre-processing.} 
We normalize the input skeleton by transforming the coordinates such that the joint position of the middle spine is the origin, the shoulder blades are parallel to the X-axis and the spine to the Y-axis, similar to Shahroudy et al.~\cite{DBLP:conf/cvpr/ShahroudyLNW16}. 
We follow the \twoSAGCN\/ pre-processing approach and divide a segment $\mathbf{X}_s$ into contiguous, non-overlapping $5$ sec.~chunks, $\mathbf{X}_s^i$, at $30$fps, i.e., $\mathbf{X}_s = (\mathbf{X}_s^1, \cdots, \mathbf{X}_s^K) $. Note that the number of chunks per segment, $K = \lceil \frac{te}{5*30} \rceil$. 
If the $K$th chunk $\mathbf{X}_s^K$ has duration $< 5$ sec., we repeat $\mathbf{X}_s^K$, and truncate at $5$ sec. 

Thus, a single sample in action recognition is a $5$ sec. motion chunk $\mathbf{X}_s^i \in \mathbf{X}_s$, and the action category labeled for its corresponding segment $y \in \mathcal{Y}_s$ in \babel.

\noindent
\paragraph{\babel~action recognition splits.} 
\babel~contains $260$ action categories with a long-tailed distribution of samples per class, unlike other popular 3D action recognition datasets \ntu~\cite{DBLP:journals/pami/LiuSPWDK20,DBLP:conf/cvpr/ShahroudyLNW16}. 
To understand the challenge posed by the long-tailed distribution of action categories in \babel, we perform experiments with 2 different datasets containing $60$ and $120$ action categories (see Table \ref{tab:act_recog}). These are motion annotation pairs of the $60$ and $120$ action categories that are obtained from the dense subset of \babel, containing $10892$ sequences described in section \ref{sec:stage_2_interface}.
While \babel-60 is already long-tailed, \babel-120 contains both extremely frequent and extremely rare classes. 
We randomly split the $13220$ sequences in \babel~into train ($60\%$), val.~($20\%$) and test ($20\%$) sets. 
We choose the model with best performance on the val.~set, and report performance on the test set. 
We provide the precise distribution of action categories for the train and val.~splits in the project webpage.

\noindent
\paragraph{Metrics.} 
\texttt{Top-1} measures the accuracy of the highest-scoring prediction. 
\texttt{Top-5} evaluates whether the ground-truth category is present among the top $5$ highest-scoring predictions. It accounts for labeling noise and inherent label ambiguity. 
Note that it also accounts for the possible presence of multiple action categories $\mathcal{Y}_s$, per input movement sequence. 
Ideal models will score all the categories relevant to a sample highly. 
\texttt{Top-1-norm} is the mean \texttt{Top-1} across categories. The magnitude of (\texttt{Top-1-norm}) - (\texttt{Top-1}) illustrates the class-specific bias in the model performance. In \babel, it reflects the impact of class imbalance on learning.

\noindent
\paragraph{Training.} We experiment with two losses -- standard cross-entropy loss, and the recently introduced focal loss \cite{DBLP:journals/pami/LinGGHD20}. 
Focal loss compensates for class imbalance by weighting the cross-entropy loss higher for inaccurate predictions. 
We observe that a class-balanced loss \cite{cui2019class} further improves performance. We refer to this setting of the class-balanced focal loss as \texttt{Focal} in Table~\ref{tab:act_recog}. 

We use the Adam optimizer \cite{KingmaB14} with a learning rate of $0.001$, and an annealing scheme that decreases the learning rate by a factor of 10 at epochs $20$, $40$, and $60$, following Shi et al.~\cite{shi2019two}. We used `Weights \& Biases'\footnote{https://wandb.ai} for tracking experiments \cite{wandb}. 

\begin{table}[t!]
    \centering
    \begin{tabular}{c|c|c|c|c}
        \hline
         \# actions & Loss type & Top-5 & Top-1 & Top-1-norm \\
        \hline
        \multirow{2}{1em}{60} &  \texttt{CE} & 73.18 & 41.14 & 24.46 \\
         &  \texttt{Focal} & 67.83 & 33.41 & 30.42 \\        
        \hline         
        \multirow{2}{1em}{120} &  \texttt{CE} & 70.49 & 38.41	& 17.56 \\
         &  \texttt{Focal} & 57.96 & 27.91	& 26.17 \\        
        \hline
    \end{tabular}
    \vspace{5pt}
    \caption{
        3D action recognition performance on different subsets of \babel~with \twoSAGCN~\cite{shi2019two}. 
        \texttt{CE} indicates Cross-Entropy loss and \texttt{Focal} indicates the combination of class-balanced \cite{cui2019class} focal loss \cite{DBLP:journals/pami/LinGGHD20}. 
    }
    \label{tab:act_recog}
\end{table}

\noindent
\paragraph{Results.} 
We observe that the decrease in \texttt{Top-1} and \texttt{Top-5} performance with the increase in number of classes is relatively small, in Table~\ref{tab:act_recog}. 
Importantly, we note that \texttt{Top-1-norm} is much lower than \texttt{Top-1}. This clearly points to inefficient learning from the long-tailed class distribution in \babel. 
The \texttt{Focal} losses significantly improves \texttt{Top-1-norm} performance on all \babel~subsets. This is encouraging for efforts to learn models with lower class-specific biases despite severe class imbalance. 

\noindent
\paragraph{\babel~as a recognition benchmark.} 
On the widely used \ntu~benchmark, \texttt{Top-1} recognition performance approaches $87\%$ with \twoSAGCN~\cite{shi2019two}. 
Note that unlike \ntu~whose distribution of motions and actions is carefully controlled, the diversity and long-tailed distribution of samples in \babel~makes the recognition task more challenging. 
The few-shot recognition split of \ntu~120 partly addresses this issue. 
However, considering few-shot learning as a separate task typically involves measuring performance on only the few-shot classes, ignoring the larger distribution. 
Models in the real world ideally need to learn and perform well on both the frequent and infrequent classes of an imbalanced distribution \cite{DBLP:conf/cvpr/WertheimerH19}. 
We present \babel~as an additional benchmark for 3D action recognition, which evaluates the ability of models to learn from more realistic distributions of actions. 

%% file: arxiv-sections/conclusion.tex
\section{Conclusion}

We presented \babel, a large-scale dataset with dense action labels for mocap sequences. 
Unlike existing 3D datasets with action labels, \babel~has labels for \emph{all} actions that occur in the sequence including simultaneously occurring actions, and transitions between actions. 
We analyzed the relationships between temporally adjacent actions and simultaneous actions occurring in a sequence. 
We demonstrated that the action recognition task on \babel~is challenging due to the label diversity and long-tailed distribution of samples. We believe that \babel~will serve as a useful additional benchmark for action recognition since it evaluates the ability to model realistic distributions of data. 
We hope that this large-scale, high quality dataset will accelerate progress in the challenging problem of understanding human movement in semantic terms. 

\paragraph{Acknowledgements.} We thank Joachim Tesch for support with rendering mocap sequences, the Software Workshop at MPI-IS for their support with the BABEL Action Recognition Challenge, Taylor McConnell, Leyre Sánchez Vinuela, Tsvetelina Alexiadis, Mila Gorecki for their support with categorization of labels, Muhammed Kocabaş for the interesting discussions, Cornelia Köhler and Omid Taheri for feedback on the manuscript. Nikos Athanasiou acknowledges funding by Max Planck Graduate Center for Computer and Information Science Doctoral Program.

\paragraph{Disclosure:} 
MJB has received research gift funds from Adobe, Intel, Nvidia, Facebook, and Amazon. While MJB is a part-time employee of Amazon, his research was performed solely at, and funded solely by, Max Planck. MJB has financial interests in Amazon, Datagen Technologies, and Meshcapade GmbH.